\documentclass[sigconf]{acmart}
\usepackage{multirow}
\usepackage{hyperref}
\usepackage{threeparttable}
\usepackage[capitalize]{cleveref}
\usepackage[utf8]{inputenc}
\usepackage{textgreek}
\usepackage{float}
\AtBeginDocument{%
  }

\setcopyright{acmlicensed}
\copyrightyear{2026}
\acmYear{2026}
\acmDOI{XXXXXXX.XXXXXXX}
\acmConference[WWW '26]{The Web Conference 2026}{April 13–April 17, 2026}{Dubai}

\acmISBN{979-8-4007-2307-0/2026/04}

\begin{document}

\title{Causal Pre-training Under the Fairness Lens: An Empirical Study of TabPFN}


\author{Qinyi Liu}
\affiliation{
  \institution{University of Bergen}
  \department{Centre for the Science of Learning \& Technology (SLATE)}
  \city{Bergen}
  \country{Norway}
}
\email{qinyi.liu@uib.no}

\author{Mohammad Khalil}
\affiliation{
  \institution{University of Bergen}
   \department{Centre for the Science of Learning \& Technology (SLATE)}
  \city{Bergen}
  \country{Norway}
}
\email{mohammad.khalil@uib.no}

\author{Naman Goel}
\affiliation{
    \department{Department of Computer Science}
  \institution{University of Oxford and Alan Turing Institute}
  \city{Oxford}
  \country{United Kingdom}
}
\email{naman.goel@cs.ox.ac.uk}

\renewcommand{\shortauthors}{Liu et al.}

\begin{abstract}
Foundation models for tabular data, such as the Tabular Prior-data Fitted Network (TabPFN), are pre-trained on a massive number of synthetic datasets generated by structural causal models (SCM). They leverage in-context learning to offer high predictive accuracy in real-world tasks. However, the fairness properties of these foundational models, which incorporate ideas from causal reasoning during pre-training, remain underexplored. In this work, we conduct a comprehensive empirical evaluation of TabPFN and its fine-tuned variants, assessing predictive performance, fairness, and robustness across varying dataset sizes and distributional shifts. Our results reveal that while TabPFN achieves stronger predictive accuracy compared to baselines and exhibits robustness to spurious correlations, improvements in fairness are moderate and inconsistent, particularly under missing-not-at-random (MNAR) covariate shifts. These findings suggest that the causal pre-training in TabPFN is helpful but insufficient for algorithmic fairness, highlighting implications for deploying TabPFN (and similar) models in practice and the need for further fairness interventions.
\end{abstract}





\maketitle

\section{Introduction}
Algorithmic fairness is a critical concern in the deployment of machine learning (ML) models for high-stakes decision-making. This concern is often compounded by issues such as spurious correlations and distribution shifts in the data~\cite{10.24963/ijcai.2024/909}. Foundation models for tabular data, such as  Tabular Prior-data Fitted Network (TabPFN)~\cite{hollmann2025tabpfn}, are pre-trained on millions of synthetic datasets based on structured causal models (SCM) and leverage in-context learning (ICL) to make accurate predictions on real datasets. This reduced reliance on real and biased data for prediction and the use of SCM generated synthetic data in pre-training motivates a systematic empirical evaluation of its fairness implications, since bias in machine learning models is partially attributed to different types of biases in the training datasets~\cite{mehrabi2021survey}
 ~(rigorous justification for the motivation is presented in \cref{sec:bg-tabpfn}).

However, despite impressive predictive accuracy and thus, growing interest in its widespread use in practical applications, the behavior of TabPFN with respect to fairness, especially under distributional shifts, has not been thoroughly studied.

We conduct a systematic empirical evaluation of TabPFN and its fine-tuned variants, which yields a comprehensive understanding of their predictive accuracy, fairness, and robustness under varying ICL dataset sizes and distribution shifts. We thus also explore the intrinsic fairness potential of SCM-based priors, revealing both strengths—such as robustness to spurious correlations—and limitations, particularly under missing-not-at-random (MNAR) covariate shifts. Our results offer practical insights for deploying these causally-informed foundation models in practice and highlighting scenarios where additional fairness intervention may be necessary.

\section{Background and Related Work}
\subsection{Fairness and its Challenge in ML}
\label{sec: fairness-challenge}
Algorithmic fairness is a cornerstone of responsible ML and the broader WWW community, ensuring that predictions do not discriminate against protected groups \cite{kouw2018introduction}. Two prevalent notions of group fairness are \textit{Demographic Parity (DP)} and \textit{Equalized Odds (EO)} \cite{hardt2016equality, wang2021analyzing}. DP requires predictions ($\hat{Y}$) to be independent of sensitive attributes ($S$): $P(\hat{Y}=1 \mid S=0) = P(\hat{Y}=1 \mid S=1)$. EO demands conditional independence given the true label ($Y$): $P(\hat{Y}=1 \mid Y=y, S=0) = P(\hat{Y}=1 \mid Y=y, S=1)$ for $y \in \{0,1\}$. Except in trivial cases (e.g., equal base rates), different fairness and accuracy criteria cannot be satisfied simultaneously, this is the fundamental challenge in algorithmic fairness~\cite{chouldechova2017fair,kleinberg2016inherent}.

Further, achieving fairness criteria is challenging in practice due to various types of biases in training datasets. Traditional ML models (e.g., logistic regression) achieve high accuracy by exploiting statistical correlations, but are vulnerable to two interrelated failures (among others), rooted in biased training data \cite{mehrabi2021survey}:
    
\noindent \textbf{1. Spurious correlations} arise from underspecified or imbalanced datasets \cite{ye2024spurious}. When data are limited, multiple hypotheses can fit the evidence equally. Following Occam’s Razor, models tend to prefer simpler hypotheses, which in biased datasets may correspond to non-causal shortcuts. For example, a model may use correlated proxies such as race or ZIP code to predict income, even in the absence of a causal link. Sampling noise and majority-group overrepresentation further reinforce such patterns, leading models to misattribute predictive power to sensitive attributes \cite{mehrabi2021survey}.
    
\noindent \textbf{2. Covariate shift} can originate from selection bias during data collection, where the distribution of features in the training set does not represent the true population. Formally, this occurs when $P_{\text{train}}(X) \neq P_{\text{test}}(X)$, often due to biased inclusion probabilities $P(\text{included} \mid X) \neq P(\text{included})$ \cite{kouw2018introduction}. As a result, models learn decision boundaries that rely on non-stable correlations specific to the training context rather than invariant causal relationships.

Both phenomena share similar origin: selection-biased data that fail to capture invariant causal mechanisms.

A promising direction for addressing the aforementioned challenges involves integrating causal reasoning into tabular models \cite{loftus2018causal, barocas2023fairness}. Rather than solely relying on correlations found in potentially biased empirical data, these approaches aim to learn stable relationships that align with underlying causal mechanisms.

\subsection{Pre-Trained Tabular Foundation Model}\label{sec:bg-tabpfn}
TabPFN is a recently proposed transformer-based foundation model for tabular supervised learning \cite{
hollmanntabpfn2023,hollmann2025tabpfn,helli2024drift}. Unlike conventional classifiers trained from scratch on individual datasets, TabPFN is \textit{pre-trained once} on millions of synthetic tabular tasks generated from SCMs. Synthetic tasks reflect different prediction tasks (diverse causal structures, noise levels, feature imbalances, etc). Rather than learning dataset-specific correlations, TabPFN approximates the posterior predictive distribution, effectively performing Bayesian inference over plausible causal mechanisms. The models use real task-specific data only at inference time, similar to ICL in LLMs~\cite{brown2020language,kenfack2025fair}.

The TabPFN design yields several theoretical advantages that can be relevant to fairness and robustness challenges discussed in \cref{sec: fairness-challenge}. First, the SCM-based prior grounds the model in stable causal dependencies, potentially mitigating reliance on spurious correlations tied to sensitive attributes in real datasets. Second, by meta-learning across a wide range of causal environments, TabPFN can potentially improve generalization under covariate and selection shifts. Last but not least, its Bayesian-style inference can potentially provide regularization against overfitting to biased or undersampled data, enhancing fairness in low-sample regimes.

Prior TabPFN studies have demonstrated state-of-the-art accuracy on a wide range of small- to medium-sized tabular benchmarks (smaller than 10,000 training samples), outperforming carefully tuned tree ensembles and AutoML systems \cite{hollmann2025tabpfn}. While these results highlight strong predictive performance, a more detailed evaluation of TabPFN's fairness, particularly under spurious correlations and covariate shift is still lacking, extending beyond initial fairness explorations in tabular in-context learning (e.g., ~\cite{kenfack2025fair}).

FairPFN has recently been proposed to directly optimize causal fairness metrics (e.g. counterfactual fairness) during the model adaptation process \cite{robertson2025fairpfn}. However, its current formulation is restricted to binary sensitive attributes. In this study, we \textit{instead} use TabPFN without such specific fairness adaptations, allowing us to study the group fairness behavior of the more widely used models (i.e base TabPFN and fine-tuned FT-TabPFN) under different stress-testing scenarios. This also allows us to study how far causal pre-training alone can improve group fairness, without further fairness interventions.

This paper, thus, addresses the following research questions (RQs):  
\begin{enumerate}
    \item[RQ1:] How do TabPFN and FT-TabPFN perform in terms of acccuracy and fairness across different dataset sizes, compared with traditional models?
    \item[RQ2:] How robustly do TabPFN and FT-TabPFN preserve both accuracy and fairness in the presence of covariate shift and spurious correlation?
\end{enumerate}

\section{Experiments}
\textbf{Datasets.} We use four standard tabular fairness benchmark datasets: Heart (303 samples, 13 features; sensitive: age, sex) \cite{uci-heart}, Bank (5,000, 14; education, family) \cite{moro2014bank}, Law (21k, 8; race, sex) \cite{wightman1998lsac}, and Adult (48k, 14; race, sex) \cite{uci-adult}. Features were standardized and categorical features one-hot encoded; all models share the same preprocessing.

\noindent \textbf{Models.} We evaluate five models on tabular classification tasks, comprising three classical baselines and two variants of TabPFN:  
(1) \textit{Logistic Regression (LR)} with L2 regularization ($C{=}0.1$),  
(2) \textit{Random Forest (RF)} with 50 trees of maximum depth 5 and minimum samples per split of 10,  
(3) \textit{Multi-Layer Perceptron (MLP)} with one hidden layer of 50 units, L2 regularization ($\alpha{=}0.01$), and up to 300 training iterations,  
(4) \textit{TabPFN} in zero-shot mode with pretraining limits disabled (allowing training on datasets exceeding 10k samples), $n\_estimators{=}2$, and inference using 5000 subsampled training examples as context,  
and (5) \textit{FT-TabPFN}, fine-tuned for 10 epochs via Adam ($\mathrm{lr}{=}1{\times}10^{-5}$), meta-batch size 1, inner batch size 5000, and cross-entropy loss on batched in-context predictions, evaluated with the same 5000-sample context. Further implementation details can be found in code, available at \url{https://github.com/ql909/An-Empirical-Study-of-TabPFN}.

\subsection{RQ1}
To answer RQ1, we evaluate model performance using accuracy and two standard fairness metrics—EO and DP—implemented via FairLearn\footnote{FairLearn: A Toolkit for Assessing and Improving Fairness in AI (Bird et al., 2020; Version 0.7.0).}. Although TabPFN is designed for small- to medium-sized tabular datasets (up to roughly 10k samples) \cite{hollmann2025tabpfn}, we also examine its behavior on larger datasets. For the Adult and Law datasets, we further evaluate TabPFN on subsets of 500, 1k, and 10k samples to analyze how fairness varies with dataset size.

\subsection{RQ2}
\label{sec:exp-rq2}
\textbf{Spurious correlations} To evaluate robustness against non-causal statistical dependencies, we follow the literature on spurious correlation stress-testing~\cite{veitch2021counterfactual} and design a \textit{pseudo-correlation perturbation} experiment. The key idea is to introduce an artificial feature $Z_{\text{spur}}$ that is strongly correlated with the target label in the training set but has its correlation sign reversed in the test set, allowing us to examine reliance on shortcut correlations versus invariant causal features.

Formally, we augment each dataset with a synthetic continuous variable $Z_{\text{spur}} \sim \mathcal{N}(\mu_y, 0.5)$, where $\mu_y = 1$ if $Y=1$ and $0$ otherwise in training data, and flipped to $\mu_y = -1$ if $Y=1$ and $0$ otherwise in the perturbed test data. Models capturing spurious dependencies should degrade in predictive performance and fairness upon inversion of the $Z_{\text{spur}}$--$Y$ correlation. We measure this sensitivity via differences in accuracy and fairness between original and flipped test sets, plus the \textit{Flip Consistency Rate} (proportion of unchanged predictions). All results are averaged over five independent runs with different random seeds for statistical reliability. Intuitively, causal-generalizing models should exhibit minimal degradation in utility and fairness upon flipping $Z_{\text{spur}}$.

\textbf{Covariate shift} We study a common form of covariate shift induced by \textit{missing-not-at-random} (MNAR) selection bias, where the probability of a sample being included in the training set depends on observed covariates and the (unobserved) outcome~\cite{wang2021analyzing}. Each biased subset contains $N=500$ samples; we simulate systematic under-representation by removing 70\% of instances from outcome- and attribute-dependent subgroups: women with high cholesterol who are less willing to seek medical attention (Heart), high-income middle-aged and older individuals reluctant to disclose finances (Bank), women with long working hours who are less willing to report income (Adult), and minority students with high LSAT scores who are less likely to be tracked (Law). To isolate bias effects beyond sample size, we construct a reduced-clean subset by randomly downsampling unbiased data to match size.  Results are averaged over five independent runs with different random seeds. We evaluate whether TabPFN is less affected by these MNAR shifts, as its meta-learned prior may promote stable predictive patterns.

\section{Results and Discussion}
\begin{figure}[!htbp]
    \centering
    \includegraphics[width=\columnwidth]{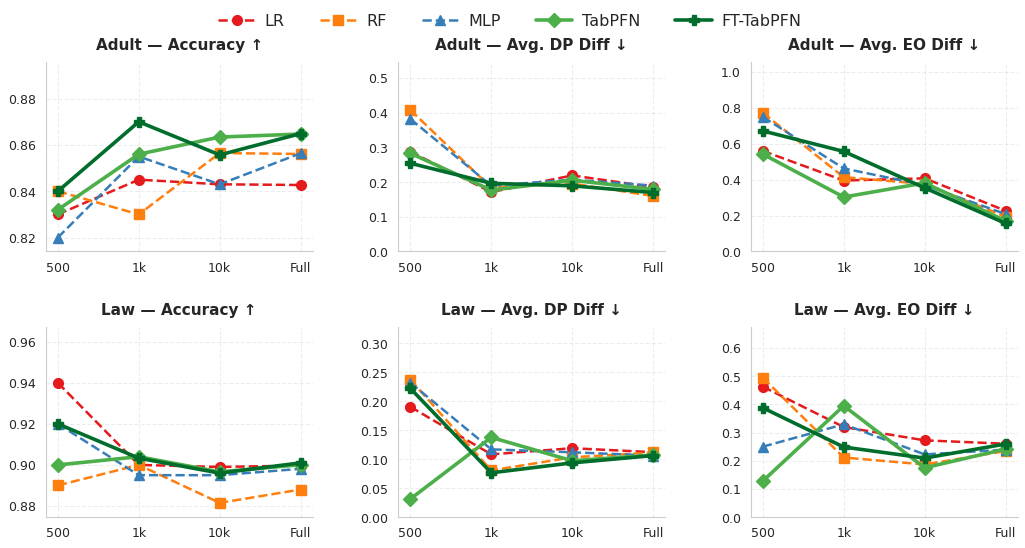}
    \vspace{-1mm}
    \caption{Performance of different models with varying dataset scale on Adult and Law datasets.}
    \label{fig:dataset_scale}
    \vspace{-2mm}
\end{figure}

\textbf{RQ1.} \cref{tab:bank_heart} reports accuracy and fairness (DP and EO w.r.t. different sensitive attributes) on Bank and Heart benchmarks. \cref{fig:dataset_scale} show results on Adult and Law benchmarks (DP, EO averaged over race and sex \footnote{The same averaging is applied to the results under covariate shift in \Cref{tab:selection_bias_results}. Complete per-attribute results are available at \url{https://github.com/ql909/An-Empirical-Study-of-TabPFN/tree/main}.}) across training set sizes ranging from 500 to full. FT-TabPFN consistently achieves the highest or near-highest accuracy across all sizes and datasets, reaching 0.99 on both Bank and Heart. TabPFN closely follows in accuracy while demonstrating superior fairness, especially in small-data regimes. For e.g, at 500 samples on Adult, average DP for TabPFN and FT-TabPFN is significantly lower than other baselines. Thus, achieving better fairness and accuracy.

As training size increases beyond TabPFN’s previously validated 10k-sample regime, both TabPFN and FT-TabPFN maintain strong accuracy. Fairness metrics, particularly EO, exhibit greater variability, showing moderate increases on Adult and slight fluctuations on Law. But the fairness performance remain at competitive levels.

\begin{table}[!htbp]
\centering
\small
\label{tab:bank_heart}
\setlength{\tabcolsep}{3pt} 
\renewcommand{\arraystretch}{0.9} 
\begin{tabular}{llccccc}
\toprule
\textbf{Dataset} & \textbf{Model} & \textbf{Acc} $\uparrow$ & \textbf{DP$_1$} $\downarrow$ & \textbf{EO$_1$} $\downarrow$ & \textbf{DP$_2$} $\downarrow$  & \textbf{EO$_2$} $\downarrow$\\
\midrule
\textbf{Bank} & LR & 0.95 & \textbf{0.10} & 0.58 & \textbf{0.05} & 0.12 \\
 & RF & \textbf{0.99} & \textbf{0.10} & 0.14 & 0.08 & 0.09 \\
 & MLP & \textbf{0.99} & 0.12 & 0.17 & 0.07 & 0.12 \\
 & TabPFN & \textbf{0.99} & \textbf{0.10} & \textbf{0.09} & 0.06 & 0.14 \\
 & FT-TabPFN & \textbf{0.99} & 0.11 & 0.10 & 0.08 & \textbf{0.08} \\
\midrule
\textbf{Heart} & LR & 0.84 & \textbf{0.23} & 0.20 & 0.20 & 0.12 \\
 & RF & 0.91 & 0.31 & 0.10 & 0.17 & 0.07 \\
 & MLP & 0.90 & 0.28 & 0.04 & 0.22 & 0.15 \\
 & TabPFN & 0.96 & 0.30 & 0.05 & \textbf{0.14} & 0.06 \\
 & FT-TabPFN & \textbf{0.99} & 0.35 & \textbf{0.03} & \textbf{0.14} & \textbf{0.03} \\
\bottomrule
\end{tabular}
{\fontsize{8}{9}\selectfont
\caption{Performance on Bank and Heart datasets. ↑ indicates the higher the better, ↓ indicates the lower the better. The same convention applies to the following tables/figures.}
}
\vspace{-1.5em}
\end{table} 

\textbf{RQ2.} \cref{tab:pseudo_flip_avg} and \cref{tab:selection_bias_results} summarize results under spurious correlations and MNAR covariate shift. As discussed in \cref{sec:exp-rq2}, a model that relies on causal mechanisms is expected to show minimal degradation when spurious attributes are flipped. Our results align with this intuition. Under pseudo-correlation perturbations, FT-TabPFN and TabPFN exhibit superior flip consistency, reaching 0.82--0.97 across datasets and leading the pack on Heart and Bank, indicating robust predictions less sensitive to the reversal of spurious attribute--label associations. RF follows as a strong second in consistency, while MLP and LR lag behind, highlighting their greater reliance on dataset-specific patterns. For Accuracy, FT-TabPFN and TabPFN have the highest Acc on most datasets (Bank 0.97, Heart/Adult 0.75-0.88). Fairness metrics (DP/EO) exhibit inconsistent patterns across models, with TabPFN/FT-TabPFN showing varying DP (Bank/Heart/Law/Adult 0.04-0.08, Adult 0.21-0.25) and EO (Heart/Law/Adult 0.09-0.12, Bank 0.22-0.27) performance.
\par Notably, the Law dataset exhibits anomalously low consistency (<0.62) and Acc (<0.56) across models post-flip, contrasting its high baseline Acc in \cref{fig:dataset_scale} (especially small samples). This flip-induced collapse persists despite strong non-perturbed utility, stemming from extreme imbalance in sensitive attributes (e.g., race/sex) and label distributions that may foster proxy reliance; with anomalies noted 
in prior work as well for this dataset~\cite{kim2025counterfactual}.
\par \cref{tab:selection_bias_results} reports model performance under MNAR-induced covariate shift. TabPFN and FT-TabPFN consistently achieve the highest accuracy on biased (Acc$_b$: 0.89--0.99) and clean (Acc$_c$: 0.86--0.99) data across datasets, outperforming baselines by 2--10\%. However, fairness improvements are inconsistent: while they attain the lowest EO in Heart and Law, EO remains notably higher in Bank (up to 0.59) and Adult shows only marginal gains. DP is particularly unstable, ranging from near-optimal in Law/Adult to the highest values in Heart/Bank (0.38--0.42).

\begin{table}[!htbp]
\centering
\small
\setlength{\tabcolsep}{3pt}
\renewcommand{\arraystretch}{0.9}
\begin{tabular}{llccccc}
\toprule
\textbf{Dataset} & \textbf{Model} & \textbf{Consistency$\uparrow$} & \textbf{Acc$\uparrow$} & \textbf{DP$\downarrow$} & \textbf{EO$\downarrow$} \\
\midrule
\textbf{Heart}
& FT-TabPFN & \textbf{0.89} & \textbf{0.88} & 0.21 & \textbf{0.10} \\
& TabPFN & 0.87 & 0.87 & 0.25 & \textbf{0.10} \\
& RF & 0.70 & 0.68 & 0.29 & 0.30 \\
& MLP & 0.56 & 0.54 & 0.10 & 0.12 \\
& LR & 0.51 & 0.49 & \textbf{0.07} & 0.11 \\
\midrule
\textbf{Bank}
& FT-TabPFN & \textbf{0.97} & \textbf{0.97} & 0.08 & 0.27 \\
& TabPFN & \textbf{0.97} & \textbf{0.97} & 0.08 & 0.22 \\
& RF & 0.96 & 0.94 & 0.07 & 0.44 \\
& MLP & 0.94 & 0.93 & 0.05 & 0.31 \\
& LR & 0.93 & 0.90 & \textbf{0.02} & \textbf{0.08} \\
\midrule
\textbf{Law}
& FT-TabPFN & 0.27 & 0.21 & 0.05 & \textbf{0.11} \\
& TabPFN & 0.27 & 0.21 & \textbf{0.04} & 0.12 \\
& RF & \textbf{0.62} & \textbf{0.56} & 0.16 & 0.18 \\
& MLP & 0.28 & 0.23 & 0.05 & \textbf{0.11} \\
& LR & 0.25 & 0.19 & \textbf{0.04} & 0.12 \\
\midrule
\textbf{Adult}
& FT-TabPFN & 0.82 & 0.76 & 0.06 & 0.10 \\
& TabPFN & 0.82 & 0.75 & \textbf{0.05} & 0.09 \\
& RF & \textbf{0.85} & \textbf{0.77} & \textbf{0.05} & 0.11 \\
& MLP & 0.81 & 0.74 & \textbf{0.05} & \textbf{0.07} \\
& LR & 0.82 & 0.74 & \textbf{0.05} & 0.08 \\
\bottomrule
\end{tabular}
\caption{Accuracy (Acc), fairness (DP / EO, averaged over sensitive attributes) under spurious correlation.}
\label{tab:pseudo_flip_avg}
\vspace{-5mm} 
\end{table}

\begin{table}[!htbp]
\centering
\small
\setlength{\tabcolsep}{2.5pt}
\renewcommand{\arraystretch}{0.9}
\begin{tabular}{llcccccc}
\toprule
\textbf{Dataset} & \textbf{Model} & \textbf{Acc$_b \uparrow$} & \textbf{Acc$_c \uparrow$} & \textbf{DP$_b \downarrow$} & \textbf{DP$_c \downarrow$} & \textbf{EO$_b \downarrow$} & \textbf{EO$_c \downarrow$} \\
\midrule
\multirow{5}{*}{\textbf{Heart}}
 & FT-TabPFN & 0.96 & 0.97 & 0.42 & 0.41 & 0.11 & 0.06 \\
 & TabPFN & \textbf{0.97} & \textbf{0.98} & 0.41 & \textbf{0.40} & \textbf{0.08} & \textbf{0.04} \\
 & RF & 0.93 & 0.95 & 0.41 & \textbf{0.40} & 0.12 & 0.07 \\
 & MLP & 0.92 & 0.94 & 0.40 & 0.42 & 0.13 & 0.16 \\
 & LR & 0.86 & 0.86 & \textbf{0.38} & 0.41 & 0.13 & 0.18 \\
\midrule
\multirow{5}{*}{\textbf{Bank}}
 & FT-TabPFN & \textbf{0.99} & \textbf{0.99} & 0.09 & 0.12 & \textbf{0.15} & 0.45 \\
 & TabPFN & \textbf{0.99} & \textbf{0.99} & 0.10 & 0.12 & 0.18 & 0.49 \\
 & RF & 0.97 & 0.98 & 0.08 & 0.11 & 0.39 & 0.50 \\
 & MLP & 0.97 & 0.98 & 0.11 & 0.12 & 0.58 & 0.59 \\
 & LR & 0.93 & 0.94 & \textbf{0.05} & \textbf{0.07} & 0.33 & \textbf{0.43} \\
\midrule
\multirow{5}{*}{\textbf{Law}}
 & FT-TabPFN & \textbf{0.89} & \textbf{0.90} & 0.15 & 0.19 & 0.33 & 0.49 \\
 & TabPFN & \textbf{0.89} & \textbf{0.90} & 0.15 & 0.19 & 0.33 & 0.49 \\
 & RF & 0.88 & 0.91 & \textbf{0.12} & \textbf{0.12} & \textbf{0.23} & \textbf{0.33} \\
 & MLP & 0.88 & 0.91 & 0.16 & 0.16 & 0.35 & 0.46 \\
 & LR & \textbf{0.89} & \textbf{0.90} & \textbf{0.12} & 0.14 & 0.27 & 0.41 \\
\midrule
\multirow{5}{*}{\textbf{Adult}}
 & FT-TabPFN & 0.81 & 0.86 & 0.23 & 0.16 & 0.34 & 0.29 \\
 & TabPFN & 0.81 & 0.86 & 0.23 & 0.16 & 0.34 & 0.29 \\
 & RF & 0.81 & 0.86 & \textbf{0.16} & \textbf{0.12} & 0.33 & 0.29 \\
 & MLP & \textbf{0.82} & \textbf{0.88} & 0.25 & 0.16 & \textbf{0.32} & \textbf{0.25} \\
 & LR & 0.81 & 0.83 & 0.21 & 0.13 & 0.37 & \textbf{0.25} \\
\bottomrule
\end{tabular}
\caption{Model performance under MNAR covariate shift. Acc$_b$, DP$_b$, EO$_b$ refer to the biased model (trained on selected data).
Acc$_c$, DP$_c$, EO$_c$ refer to the clean/unbiased baseline.}
\label{tab:selection_bias_results}
\vspace{-0.6cm}
\end{table}

\section{Conclusion}
Our empirical evaluation yields three key findings. First, the TabPFN family achieves superior predictive accuracy across dataset sizes and demonstrates robustness to spurious correlations, yet its fairness improvements are moderate and inconsistent, with group fairness metrics like EO fluctuating with scale. Second, and critically, we uncover a previously overlooked asymmetry: while causal pre-training effectively mitigates spurious statistical shortcuts, it provides limited to no fairness improvement under realistic MNAR selection biases, possibly because causal pre-training cannot compensate for the systematic under-representation of causal data from subpopulations, one of the main sources of unfairness in real-world settings. Third, this finding reveals a fundamental gap: causal pre-training emerges as a promising direction for bridging causal robustness and algorithmic fairness, yet it may fall short in entangled scenarios, such as imbalanced sensitive attributes. Future work could explore integrating causal imputation for missing attributes and developing dynamic fairness regularization.


\bibliographystyle{ACM-Reference-Format}
\bibliography{main}

\end{document}